\documentclass[sigconf]{acmart}
\AtBeginDocument{%
  }

\setcopyright{acmlicensed}
\copyrightyear{2018}
\acmYear{2018}
\acmDOI{XXXXXXX.XXXXXXX}
\acmConference[Conference acronym 'XX]{Make sure to enter the correct
  conference title from your rights confirmation email}{June 03--05,
  2018}{Woodstock, NY}
\acmISBN{978-1-4503-XXXX-X/2018/06}

\usepackage{multirow} 
\usepackage{amsmath}
\usepackage{bm}

\begin{document}

\title{Why Retrieval-Augmented Generation Fails: A Graph Perspective}

\author{Kai Guo}
\affiliation{%
  \institution{Michigan State University}
  \country{}
}
\email{guokai1@msu.edu}

\author{Xinnan Dai}
\affiliation{%
  \institution{Michigan State University}
  \country{}
}
\email{daixinna@@msu.edu}

\author{Zhibo Zhang}
\affiliation{%
  \institution{Michigan State University}
  \country{}
}
\email{zhan2185@msu.edu}

\author{Nuohan Lin}
\affiliation{%
  \institution{Michigan State University}
  \country{}
}
\email{linnuoha@msu.edu}

\author{Shenglai Zeng}
\affiliation{
  \institution{Michigan State University}
  \country{}
}
\email{zengshe1@msu.edu}

\author{Jie Ren}
\affiliation{
  \institution{Massachusetts Institute of Technology}
  \country{}
}
\email{rjie@mit.edu}

\author{Haoyu Han}
\affiliation{
  \institution{Michigan State University}
  \country{}
}
\email{hanhaoy1@msu.edu}

\author{Jiliang Tang}
\affiliation{
  \institution{Michigan State University}
  \country{}
}
\email{tangjili@msu.edu}

\renewcommand{\shortauthors}{Trovato et al.}

\begin{abstract}
Retrieval-Augmented Generation (RAG) has become a powerful and widely used approach for improving large language models by grounding generation in retrieved evidence. However, RAG systems still produce incorrect answers in many cases. Why RAG fails despite having access to external information remains poorly understood.
We present a model-internal study of retrieval-augmented generation that examines how retrieved evidence influences answer generation. Using circuit tracing, we construct attribution graphs that model the flow of information through transformer layers during decoding. These graphs represent interactions among retrieved context, intermediate model activations, and generated tokens, providing a graph, circuit-level view of how external evidence is integrated into the model’s reasoning process across multiple question answering benchmarks, we observe consistent structural differences: correct predictions exhibit deeper reasoning paths, more distributed evidence flow, and a more structured pattern of local connectivity, while failed predictions show shallower, fragmented, and overly concentrated evidence flow.
Building on these findings, we develop a graph-based error detection framework that uses attribution-graph topology features. Furthermore, we show that attribution graphs enable targeted interventions. By reinforcing question-constrained evidence grounding, we reshape internal routing so that answer generation remains guided by the question, leading to more effective integration of retrieved information and fewer errors.

\end{abstract}


\received{20 February 2007}
\received[revised]{12 March 2009}
\received[accepted]{5 June 2009}

\newcommand{\jt}[1]{{\color{red}[JT:#1]}}
\maketitle

\section{Introduction}

Retrieval-Augmented Generation (RAG) has become a central paradigm for improving large language models by grounding generation in external evidence~\cite{lewis2020retrieval,gao2023retrieval,han2024retrieval,chen2024benchmarking,zheng2025retrieval,su2025parametric}. By retrieving relevant documents at inference time and conditioning the model on this information, RAG systems aim to reduce incorrect predictions and improve factual reliability~\cite{ayala2024reducing,hu2025removal,niu2024ragtruth,peng2025graph,asai2023self}. Despite these advantages, incorrect outputs remain common even when the retrieved passages contain the necessary evidence. This suggests that the presence of evidence alone does not guarantee that it is faithfully integrated into the model’s reasoning process~\cite{guo2025empowering,gupta2024comprehensive,zhou2024trustworthiness,wang2023learning,shao2023enhancing}.

Existing work to investigate RAG failures focuses primarily on retrieval quality or consistency at the output-level~\cite{trivedi2023interleaving,edge2024local}. Some methods improve retrievers or re-rank retrieved documents, while others detect errors using answer–document overlap or model confidence~\cite{yu2024rankrag,lee2025shifting,wu2025multirag}. Although these approaches provide useful diagnostic indicators, they offer limited insight into the model-internal reasoning dynamics that lead to unfaithful generation.  
Recent studies have explored hidden-state representations as diagnostic signals for knowledge checking~\cite{zeng2025towards}. However, such approaches typically rely on representations from a single layer and only provide a largely static view of the model’s internal state~\cite{liu2025selfelicit}. As a result, they do not characterize how the retrieved evidence is propagated, transformed, and combined across layers during decoding. This highlights the need for a methodological framework that explicitly captures internal evidence flow, enabling a granular understanding of knowledge aggregation.

In this work, we take a graph perspective on RAG reasoning. Instead of examining only inputs and outputs, we analyze how retrieved evidence propagates through the model during decoding. 
We utilize the circuit tracing technique ~\cite{ameisen2025circuit} to build topological features to quantify how context tokens influence intermediate activations and final answer tokens.
We then translate these attribution signals into attribution graphs, which represent information flow among retrieved tokens, intermediate components, and generated outputs. This graph-based representation enables us to perform direct structural analysis of reasoning processes across examples. Therefore we conduct a systematic study of both correct and incorrect RAG predictions. We observe consistent structural differences across datasets.  
Correct predictions exhibit deeper reasoning paths, more distributed evidence flow, and a more structured local connectivity. In contrast, incorrect predictions show shallower, fragmented, and overly concentrated evidence flow.

To further provide a clear explanation of why failures occur, we focus on a mixed-context setting in which retrieved passages contain both supporting and distracting information. This scenario is particularly diagnostic, as successful reasoning requires selectively integrating the truly relevant evidence rather than relying on superficial question–context overlap.
Tracing internal information flow under this condition reveals a recurring failure mode that we term \textbf{surface-aligned evidence grounding (SAEG)}: evidence only superficially matches the question but lacks deep understanding of the question and sustained influence from it, while generation becomes increasingly dominated by retrieved context.  
In contrast, correct predictions often exhibit \textbf{question-constrained evidence grounding (QCEG)}, where the model places stronger emphasis on understanding the question and retrieved evidence remains consistently regulated by the question’s semantic constraints, forming deeper and more integrated reasoning structures.

Overall, our study establishes attribution-graph structure as a practical and interpretable lens for understanding evidence-grounding failures in RAG systems. Building on the above insights, we develop model-internal error detection methods and targeted inference-time interventions that directly regulate internal routing dynamics. These approaches not only detect incorrect predictions but can also steer some failures toward correct outcomes, demonstrating the practical utility of our mechanistic understanding.
Our main contributions are summarized as follows.
\begin{itemize}
    \item We use circuit tracing to derive attribution graphs for RAG models, enabling a graph-based analysis of evidence propagation and influence.
    \item We identify consistent structural differences between correct and incorrect predictions, showing that many RAG errors stem from insufficient question understanding and over-reliance on retrieved context.
    \item We develop a graph-based error detection framework that operates purely on internal model dynamics.
    \item We demonstrate that attribution-graph analysis enables targeted inference-time interventions that promote \textbf{question-constrained evidence grounding (QCEG)}, thereby reducing incorrect predictions during generation.
\end{itemize}

\section{Related Work}
Due to space limitations, we provide a brief overview of the most relevant prior work here and defer a more comprehensive discussion to the Appendix~\ref{sec:related}.

\textbf{Retrieval-Augmented Generation.}
Retrieval-Augmented Generation (RAG) improves the factuality and reasoning of large language models by grounding generation in external knowledge~\cite{zhao2026retrieval,fan2024survey}. Prior work has explored dense and hybrid retrieval, multi-hop evidence gathering, iterative retrieval–generation loops, and query reformulation~\cite{nian2025w,tang2024multihop,trivedi2023interleaving,chan2024rq}. Other efforts enhance robustness through context selection, reranking, compression, and prompt engineering~\cite{dong2024don,ampazis2024improving}.  

Despite these advances, most approaches treat the language model as a black box and focus on system-level improvements, offering limited insight into how retrieved evidence is internally processed. Consequently, they cannot fully explain why errors persist even when relevant evidence is successfully retrieved. Some recent work evaluates faithfulness and evidence usage~\cite{zeng2025towards,liu2025selfelicit}, but typically relies on representations from a single layer, providing only a static and partial view of the model’s internal computation.

\textbf{Interpretability and Circuit Analysis of LLMs.}
A parallel line of research investigates transformer internals using attention analysis, Sparse Autoencoders, transcoders, and circuit tracing~\cite{clark2019does,vig2019analyzing,cunningham2023sparse,dunefsky2024transcoders,elhage2021mathematical}. These methods decompose neural representations into interpretable components and reveal that specific behaviors can often be attributed to distributed circuits spanning layers and heads~\cite{paulo2024automatically,ferrando2024know}. Attribution graphs have emerged as a useful abstraction for modeling information flow within networks~\cite{marks2024sparse}.

Although circuit-level analyses have shed light on reasoning in standalone language models~\cite{dai2025graphghost,zhao2025verifying}, they rarely consider retrieval-augmented settings. As a result, how externally retrieved evidence interacts with internal computational circuits in RAG remains largely unexplored.

\section{Background and Preliminaries}
\label{sec:background}

In this section, we formally define the attribution graph and describe how it is constructed from the internal computation of a transformer model.

\subsection{Definition of Attribution Graphs}
\label{sec:def_g}

We represent token-level causal interactions inside the model in a graph view. In particular, we model the interactions among the activations as a directed attribution graph $G = (V, E)$
that captures how information flows between token representations across layers during inference.

Each node \( v_{t,\ell} \in V \) corresponds to the representation of token position \(t\) at transformer layer \(\ell\). A directed edge \( (v_{s,k} \rightarrow v_{t,\ell}) \in E \) indicates that the token state at position \(s\) in layer \(k\) contributes to the token state at position \(t\) in layer \(\ell\). The edge weight \(w\) measures the strength of this causal contribution. This graph-level view allows us to analyze model reasoning as a structured computational process, revealing how evidence is integrated, propagated, and transformed as representations evolve across layers.
\subsection{Constructing Attribution Graphs}

We now describe how token-level attribution graphs are constructed from a transformer model.

\paragraph{Feature Decomposition as the Node Basis}

Following prior work on circuit tracing and attribution \cite{dunefsky2024transcoders}, we adopt transcoders to decompose residual stream activations when building the attribution graph for a fixed target logit. At each layer $\ell$ and token position $t$, the residual stream vector is represented as a sparse set of learned activation units, which serve as intermediate carriers of attribution signals.

Attribution is computed at the level of activation units, reflecting how each unit contributes—directly or indirectly—to the target logit through the network. These activation-unit-level attributions are then aggregated by token position, so that tokens in the prompt, retrieved context, and generated output correspond to nodes in the attribution graph.

\paragraph{Edge Construction via a Linearized Replacement Model}
Edge weights in the attribution graph are obtained using a locally linearized replacement model, following existing circuit-tracing methods \cite{ameisen2025circuit}. Specifically, we replace MLP blocks with their corresponding transcoders while keeping attention modules unchanged, and fix attention patterns and layer-normalization terms at their forward-pass values. Under this setting, the network computation is linear with respect to activation-unit activations.

This linearization allows attribution signals with respect to the target logit to be decomposed into additive contributions between activation units. These unit-level attributions are aggregated across units associated with each token pair, yielding directed token-to-token attribution scores, which are used as edge weights in the attribution graph.

\section{Circuit Analysis for RAG}
In this section, we analyze the internal circuit structure of RAG models to understand how retrieved evidence is integrated during answer generation. We begin with a general retrieval setting, where the retrieved context is treated as unconstrained. Under this setting, we compare the attribution-graph structures of \textbf{correct and incorrect} predictions to identify systematic differences in how information flows through the model’s internal computation.

To probe the failure mechanism more directly, we introduce a more challenging \textbf{mixed-context setting} in which retrieved passages intentionally include both supporting and non-supporting information. This scenario better reflects realistic retrieval conditions and places stronger demands on the model’s ability to distinguish relevant evidence from noise. Analyzing circuit behavior under this mixed setting allows us to study how incorrect reasoning emerges when the model fails to selectively ground its predictions in truly supportive context.
\subsection{Circuit Analysis of Correct and Incorrect
Predictions}

This section uses attribution graphs to analyze differences between correct and incorrect predictions in how models internally organize and integrate information. We examine the structural properties of the model’s internal computation during answer generation.
These structural patterns provide insight into the internal mechanisms that distinguish successful from unsuccessful prediction use.

\subsubsection{Graph Metrics}
\label{sec:graph-signatures}
\label{graph_sig}
To understand why some predictions successfully integrate retrieved evidence while others do not, we examine the structural organization of their attribution graphs. Given an attribution graph $G=(V,E)$ defined in Section~\ref{sec:def_g}, each example is summarized through a set of graph-level statistics. Rather than characterizing individual graphs separately, our goal is to identify systematic structural differences between correct and incorrect reasoning circuits.

We hypothesize that correct and incorrect predictions differ along three fundamental dimensions of internal evidence integration:  
(1) how far information propagates through the model,  
(2) how strongly token representations interact with one another, and  
(3) how information is organized across local and global structures.  
We therefore design a set of graph metrics that quantify each of these aspects.

\paragraph{Propagation depth.}

The first dimension concerns the depth of information propagation. Correct reasoning may require evidence to travel through multiple intermediate representations, whereas shallow propagation may reflect shortcut or surface-level processing. We measure this using the longest directed path length,  
$\bm{\mathrm{DAG\text{-}L}(G)} = \max_{\pi \in \mathcal{P}(G)} |\pi|$,  
where $\mathcal{P}(G)$ is the set of directed paths in $G$. Larger values indicate longer multi-step propagation chains, suggesting more compositional reasoning.

\paragraph{Interaction strength.}

The second dimension captures how strongly token representations interact during computation. If evidence is effectively integrated, we expect richer connectivity among tokens rather than isolated or weakly connected fragments.

We measure this using two complementary metrics. The average degree  
$\bm{\mathrm{AvgDeg}(G)}=\frac{1}{|V|}\sum_{v\in V}(\deg^{\text{in}}(v)+\deg^{\text{out}}(v))$  
captures the typical number of interactions per token. Here, $\deg^{\text{in}}(v)$ and $\deg^{\text{out}}(v)$ denote the in-degree and out-degree of node $v$. The directed edge density  
$\bm{\mathrm{Dens}(G)}=\frac{|E|}{|V|(|V|-1)}$  
measures how densely the reasoning circuit is connected overall. Higher values indicate stronger and more widespread evidence interaction.

\paragraph{Structural organization across scales.}

We further characterize how information is organized at both local and global scales.

Local fragmentation is captured by the fraction of disconnected triads,  
$\bm{T_{\text{disc}}(G)} = \#\text{disc}(G) \big/ \sum_{\tau}\#\tau(G)$,  
where $\#disc$ is a disconnected triad consists of three nodes with no edges among them. Larger values indicate that nearby nodes fail to interact, suggesting fragmented local structure.

Branching-style local aggregation is measured by  
$\bm{T_{\text{branch}}(G)} = \#\text{branch}(G) \big/ \sum_{\tau}\#\tau(G)$,  
where a branch triad is a three-node pattern in which two nodes both point to the same third node. Higher values indicate that information from multiple sources tends to merge into a single intermediate node, reflecting localized aggregation rather than linear propagation.

Finally, global concentration of information flow is captured by  
$\bm{\mathrm{MaxPR}(G)} = \max_{v\in V} \mathrm{PR}(v)$,  
where $\mathrm{PR}(v)$ is the PageRank score of node $v$. Larger values indicate that information flow is dominated by a single hub, whereas lower values suggest more distributed integration.

\medskip

Together, these six metrics provide a structural signature of each reasoning circuit. By comparing these signatures between correct and incorrect predictions, we can identify how successful evidence integration differs from failure at the level of internal computation.



\subsubsection{A Study of Correct and Incorrect Question Answering}

\paragraph{\textbf{Setup}}
We study QA benchmarks (HotpotQA~\cite{yang2018hotpotqa}, 2WikiMultihopQA~\cite{ho2020constructing}, and MuSiQue~\cite{trivedi2022musique}), where answering requires composing evidence across multiple passages. For each query, we retrieve a fixed-size context and generate chain-of-thought answers using LLaMA-3 8B Instruct~\cite{dubey2024llama} with greedy decoding. We then assign a binary label $y \in \{0,1\}$ (incorrect or correct) using an external LLM-based judge, Gemini-2.5-Flash-Lite~\cite{comanici2025gemini}. Finally, for each dataset, we construct a balanced set of attribution graphs comprising 500 incorrect and 500 correct predictions to enable directly comparable structural analyses across classes.

Within each dataset, we construct balanced subsets consisting of 500 correct and 500 incorrect predictions, ensuring that structural comparisons between attribution graphs are directly comparable across classes.

\paragraph{\textbf{Finding 1: Correct Answers Arise from Deeper, More Structured, and More Evenly Distributed Circuits}}
\label{sec:six}

A consistent structural contrast emerges across datasets:  
\textbf{correct reasoning circuits are deep, densely interconnected, and broadly distributed, whereas wrong circuits are shallow, sparse, and overly centralized.}

\begin{figure}     \centering     
\setlength{\abovecaptionskip}{0.cm}     \setlength{\belowcaptionskip}{0.cm}     \includegraphics[scale=0.19]{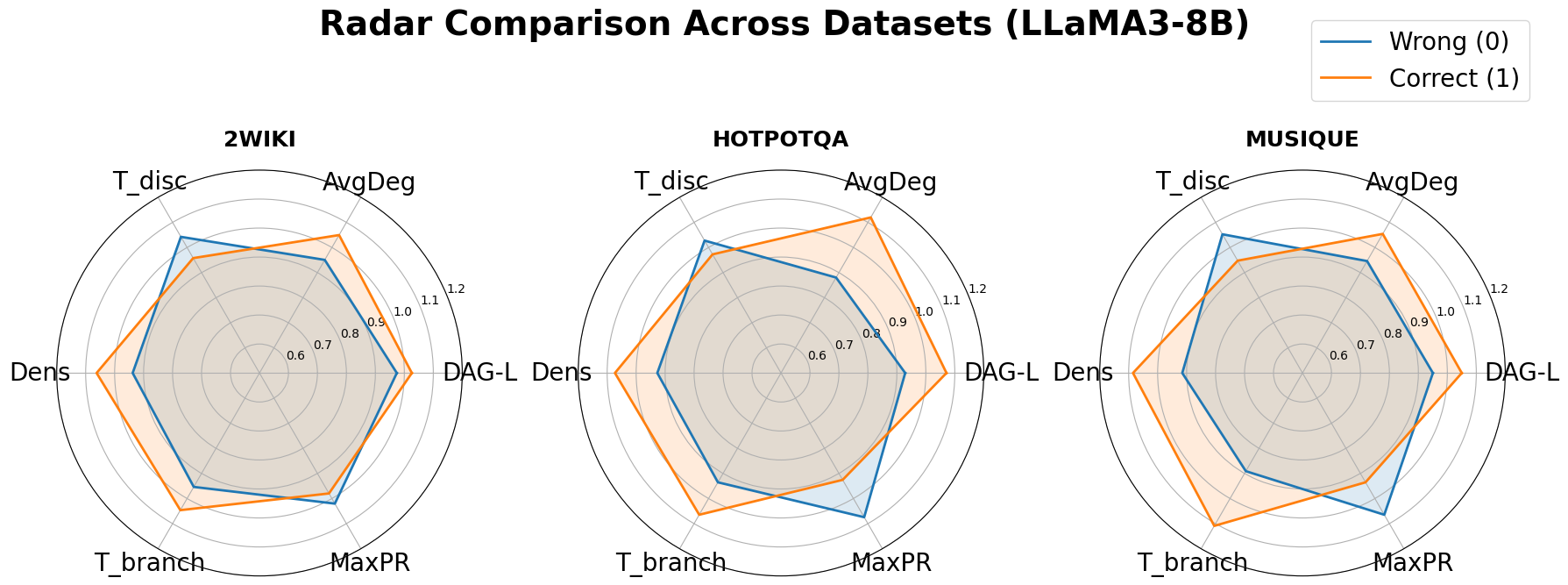}  
\captionsetup{skip=10pt}
\vskip -1em
\caption{Radar comparison of  attribution-graph structural metrics between correct and wrong predictions across three QA datasets (2Wiki, HotpotQA, and MuSiQue).}   \label{fig: radar} 
\vskip -1em
\end{figure}

Figure~\ref{fig: radar} shows this separation across all structural metrics. Correct answers are supported by attribution graphs that are deeper, more structurally organized, and more evenly distributed in how evidence is utilized. Incorrect answers, in contrast, arise from circuits that are shallow, fragmented, and overly concentrated around a few dominant nodes. In addition to LLaMA-3 8B Instruct, we also analyze Qwen-3 8B in Figure~\ref{fig:qwen} in Appendix. The results show consistent conclusions: correct and incorrect predictions exhibit clearly different structural patterns.

\textbf{Deeper vs.\ shallower propagation.}
Correct graphs exhibit longer directed propagation depth (higher DAG-L), indicating that evidence signals travel through multi-step internal routes before reaching the final answer tokens. Incorrect predictions show shorter directed paths, suggesting that reasoning is truncated and relies on fewer intermediate transformations.

\textbf{Structured vs.\ fragmented connectivity.}
Correct circuits are more structurally organized at both global and local levels. Globally, they display higher interaction richness, reflected in larger average degree (AvgDeg) and edge density (Dens), indicating stronger cross-token coupling and more integrated evidence flow. Locally, they contain fewer disconnected triads (lower $T_{disc}$), meaning that neighboring nodes are more likely to participate in coordinated interactions. Incorrect circuits, by contrast, are sparser and more fragmented, with many local node groups remaining structurally isolated. In addition, higher branching motifs (higher $T_{\text{branch}}$) indicate that intermediate states more often distribute information to multiple downstream components, reflecting a more structured and distributed reasoning process.

\textbf{Distributed vs.\ concentrated evidence flow.}
Finally, correct circuits make use of evidence in a more evenly distributed manner. They exhibit lower maximum PageRank values (MaxPR), indicating that importance is spread across multiple nodes rather than dominated by a single hub. 

Taken together, these patterns show that correct reasoning emerges from circuits that are deep, structurally coherent, and broadly distributed in their use of evidence. Incorrect answers, by contrast, arise from shallow, fragmented, and overly centralized circuits in which information either fails to propagate sufficiently or becomes concentrated on a small set of dominant nodes.

\begin{figure}     \centering     
\setlength{\abovecaptionskip}{0.cm}     \setlength{\belowcaptionskip}{0.cm}     \includegraphics[scale=0.17]{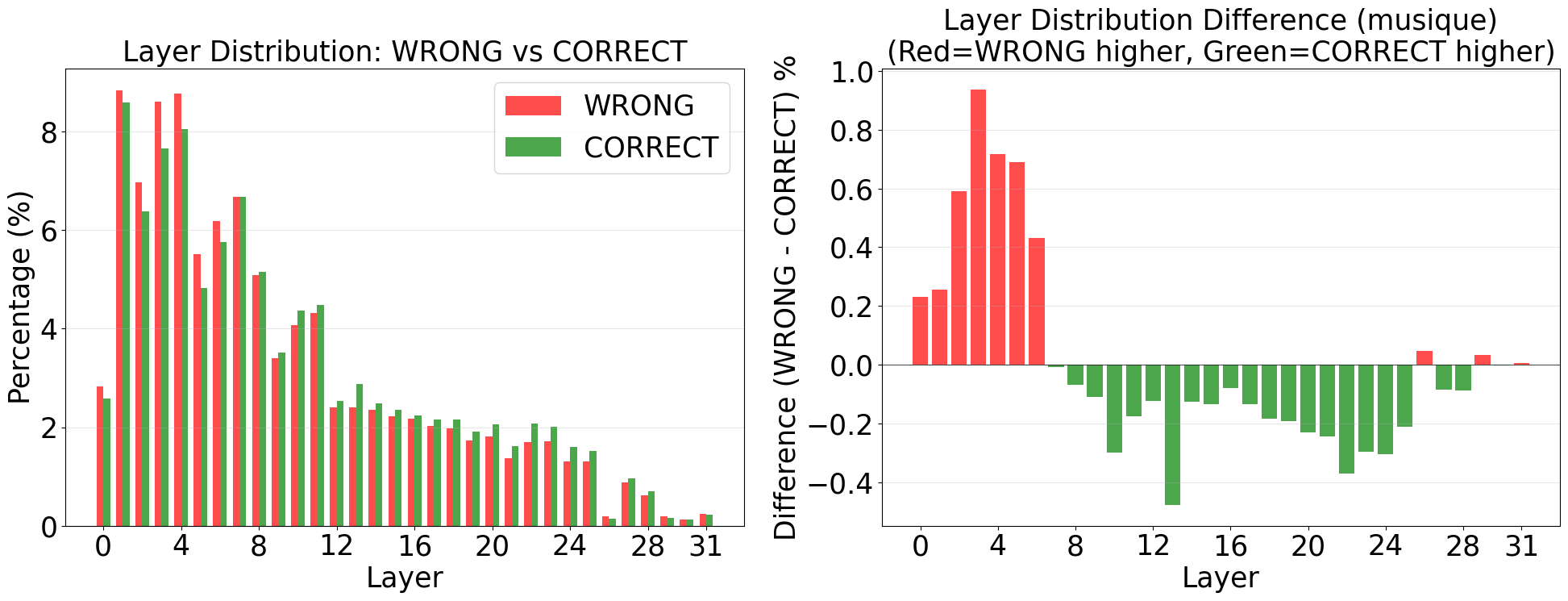}  
\captionsetup{skip=10pt}
\vskip -1em
\caption{Layer-wise attribution mass for correct and wrong predictions (left) and their difference on MuSiQue (right).
}   \label{fig: shallow} 
\vskip -0.4em
\end{figure}

\paragraph{\textbf{Finding 2: Correct Predictions Use More Mid-Layer Processing}}

In addition to graph structure, we analyze how attribution mass is distributed across transformer layers. Figure~\ref{fig: shallow} shows a clear depth shift between correct and wrong predictions on MuSique. Due to space constraints, the corresponding results for HotpotQA and 2Wiki are provided in Figure~\ref{fig: shallow_hot} and Figure~\ref{fig: shallow_2wiki} in the Appendix.

Correct predictions allocate a larger fraction of the total activated neurons to the middle layers (approximately layers 8–18), indicating greater reliance on mid-layer reasoning computations. These layers are where the model typically combines information from different tokens and builds integrated representations. The higher activity here suggests that correct answers depend on sustained internal processing that brings together evidence from the question and retrieved context.

Incorrect predictions follow a different pattern. They rely more on early layers. Higher early-layer activity indicates that the model may be matching surface-level patterns from the retrieved text without deeper integration. 

Overall, correct answers are associated with deeper and more sustained information processing, while wrong answers tend to reflect shallower  decision dynamics.

\begin{figure}     \centering     
\setlength{\abovecaptionskip}{0.cm}     \setlength{\belowcaptionskip}{0.cm}     \includegraphics[scale=0.22]{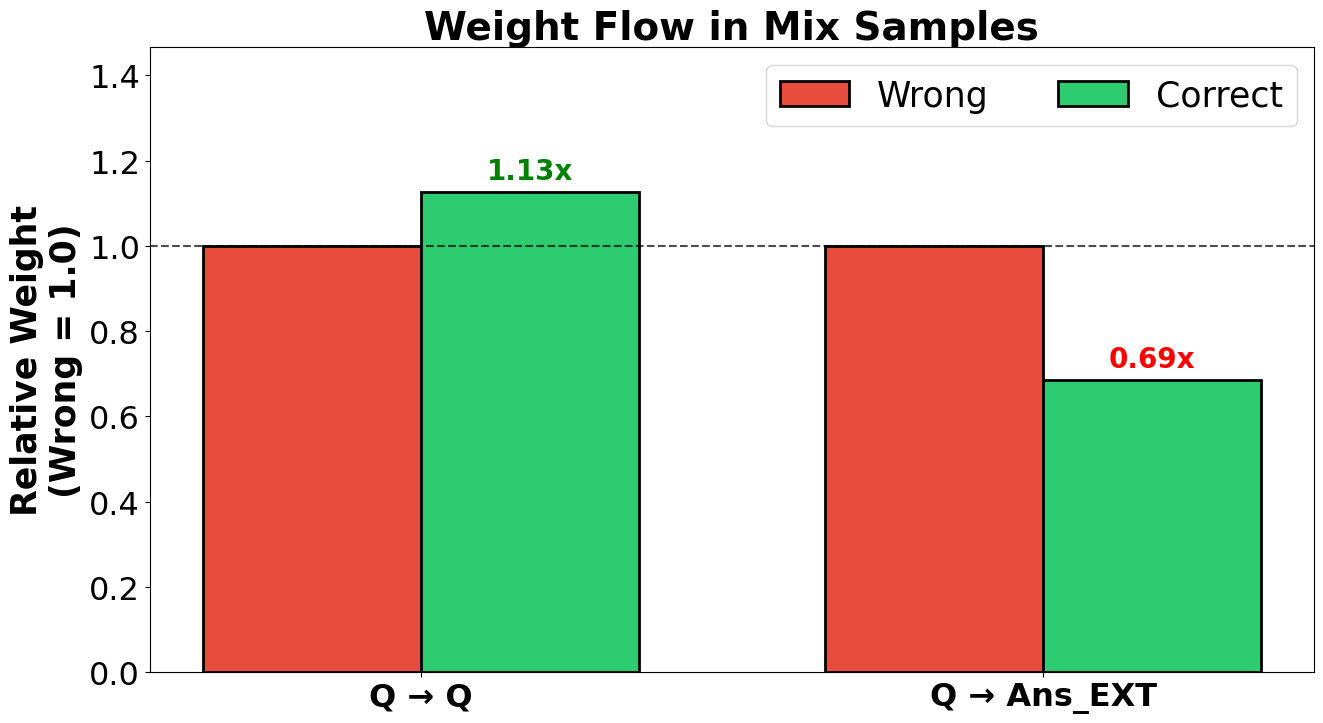}  
\captionsetup{skip=10pt}
\vskip -1em
\caption{Region-level attribution comparison between correct and wrong predictions in the mixed-context setting. Bars show relative weights for $Q \rightarrow Q$, and $Q \rightarrow \mathrm{Ans\_EXT}$.
}   \label{fig: qq0} 
\end{figure}

\subsection{Circuit Analysis under Mixed Context}
\label{sec:mix_dynamics}

The structural circuit analysis above establishes \emph{what} differentiates correct and incorrect predictions: correct reasoning is supported by deeper, more connected, and more integrative attribution graphs, whereas incorrect predictions rely on fragmented and weakly coordinated structures. We now turn to a complementary question: \emph{how do these structural differences emerge over the course of computation?}

To address this, we examine the routing dynamics of information across layers during decoding. Our analysis focuses on a mixed-context setting. In this setting, each question is paired with retrieved context that intentionally contains both supporting and non-supporting passages.

This scenario is particularly informative because the model must not only leverage external evidence, but also distinguish relevant signals from distractors while remaining aligned with the question. As a result, it provides a controlled testbed for analyzing how internal routing dynamics differ when evidence selection and integration become genuinely challenging.

We analyze these dynamics by grouping tokens into functional regions and tracking how attribution mass flows between them across layers. This provides a stage-wise view of how the model balances question understanding, reliance on externally grounded answer content, and internally composed answer representations during reasoning.

\begin{figure}     \centering     
\setlength{\abovecaptionskip}{0.cm}     \setlength{\belowcaptionskip}{0.cm}     \includegraphics[scale=0.138]{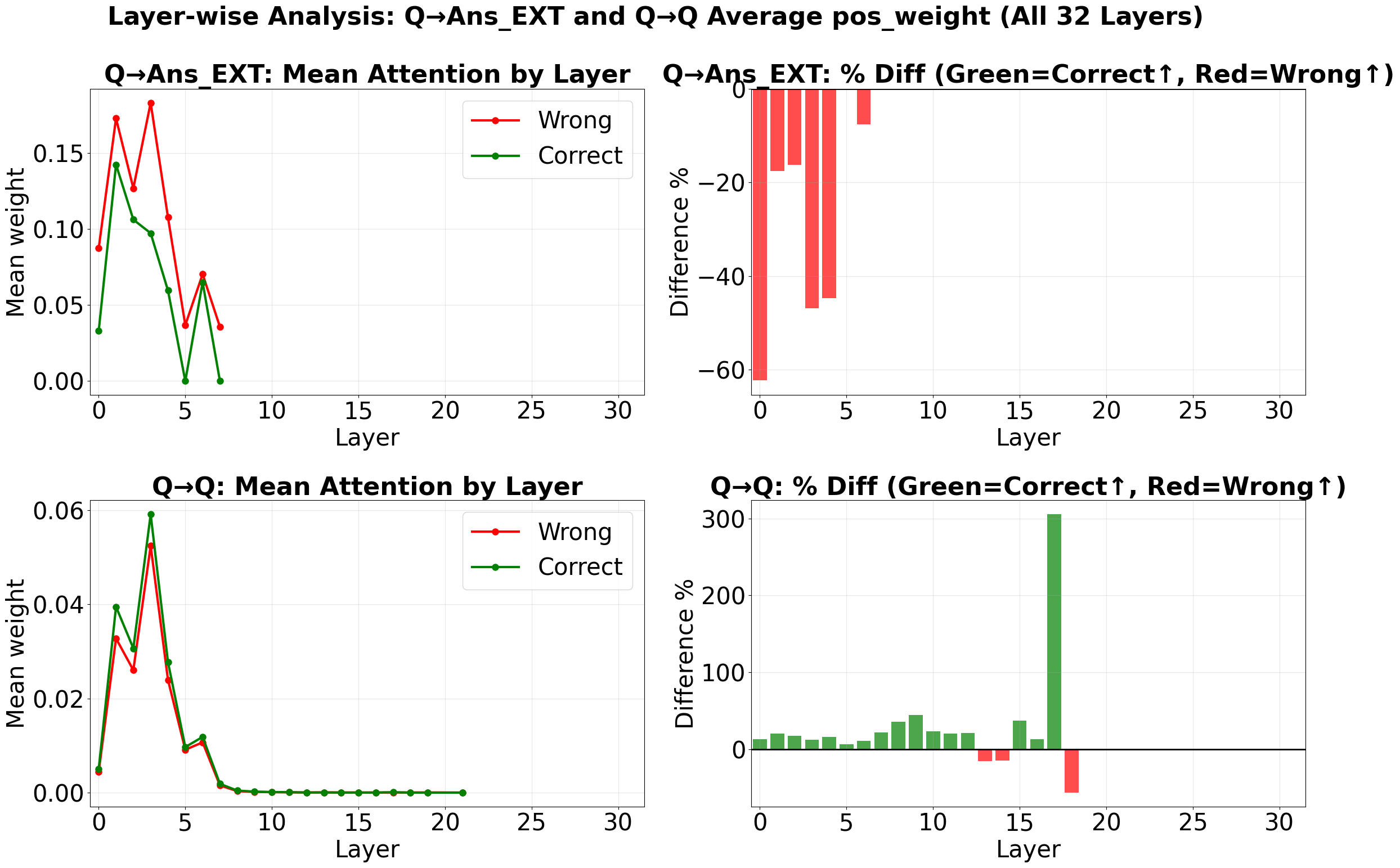}  
\captionsetup{skip=10pt}
\vskip -1em
\caption{Layer-wise attribution comparison for $Q \rightarrow \mathrm{Ans\_EXT}$ and $Q \rightarrow Q$. Left: mean routing strength per layer for correct and wrong predictions. Right: relative differences (green = correct higher, red = wrong higher).
}   \label{fig: qq} 
\vskip -0.4em
\end{figure}

\begin{figure}     \centering     
\setlength{\abovecaptionskip}{0.cm}     \setlength{\belowcaptionskip}{0.cm}     \includegraphics[scale=0.138]{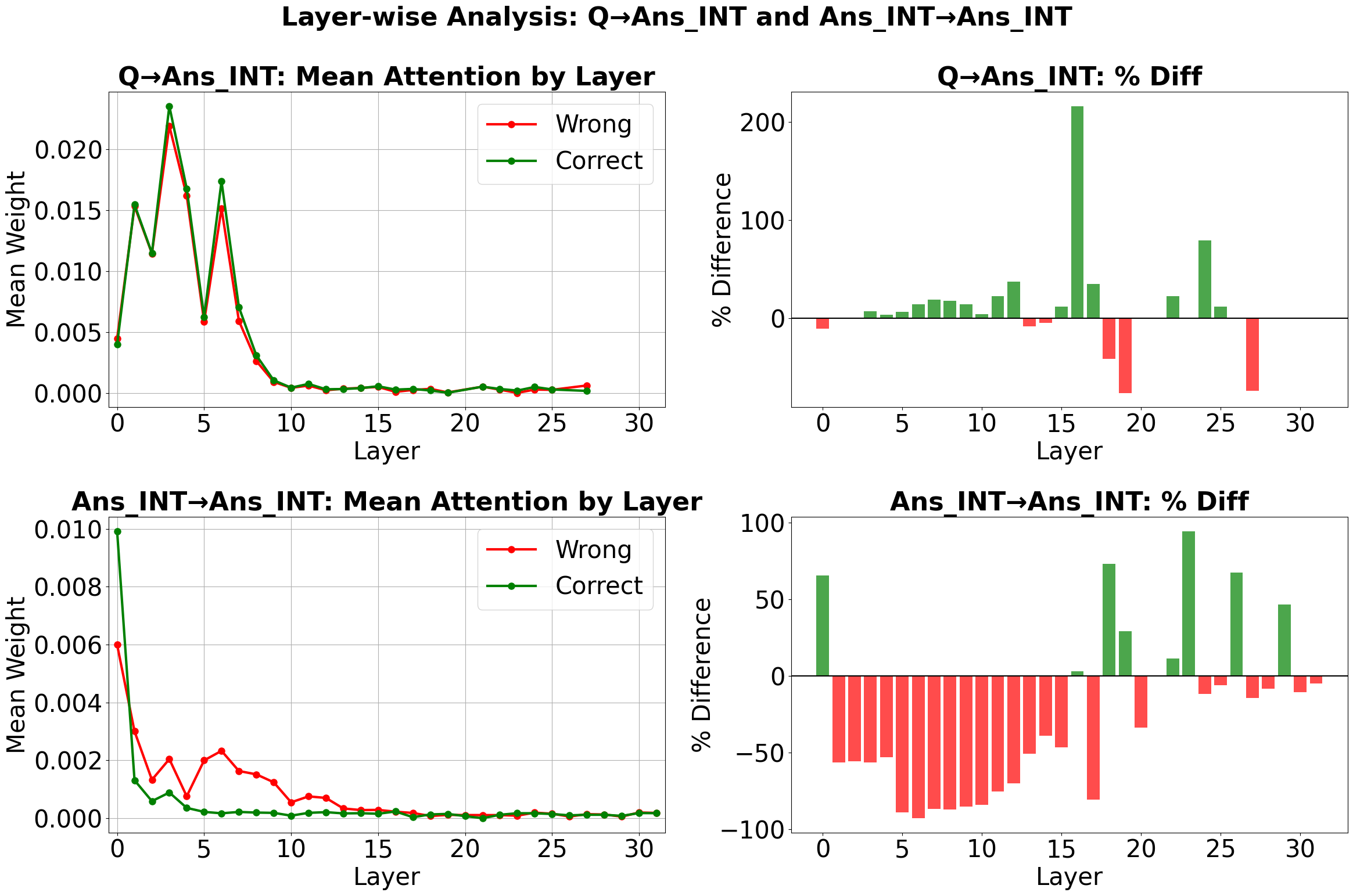}  
\captionsetup{skip=10pt}
\vskip -1em
\caption{Layer-wise attribution weight for multiple region-level edge types under mixed-context decoding, comparing correct and wrong predictions.
}   \label{fig: qin} 
\vskip -2em
\end{figure}

\subsubsection{Region-Level Routing Decomposition}

We partition tokens into three functional regions based on their roles in the reasoning process: $Q$ is the input question tokens; $\mathrm{Ans\_EXT}$ indicates answer tokens that are attributable to retrieved external context; and $\mathrm{Ans\_INT}$ denotes answer tokens that are generated internally by the model and have no direct alignment with retrieved context.

For each transformer layer $\ell$, we measure how attribution flows between these regions. Let $a(i \rightarrow j)$ denote the weight from source token $i$ to target token $j$ at layer $\ell$ 
We then aggregate attribution mass at the region level as

\begin{equation}
A^{(\ell)}_{X\rightarrow Y}
=\sum_{i \in X}\sum_{j \in Y} a(i\rightarrow j),
\qquad X,Y\in\{Q,\mathrm{Ans\_EXT},\mathrm{Ans\_INT}\},
\end{equation}

\noindent where $A^{(\ell)}_{X\rightarrow Y}$ measures the total attribution routed from region $X$ to region $Y$ at layer $\ell$ by summing over all source–target token pairs $(i,j)$ with $i \in X$ and $j \in Y$. This aggregation yields a layer-wise routing profile that reveals how the model distributes computation between understanding the question, leveraging externally aligned answer content, and developing internally composed answer representations.

\subsubsection{A Study of Reasoning Patterns in the Mixed-Context Setting}

\paragraph{\textbf{Setup}}
Our analysis centers on a mixed-context setting derived from MuSiQue, which we term \textbf{Mix-MuSiQue}. In this variant, each question is paired with a retrieved context deliberately constructed to include both supporting and non-supporting passages. This design creates a controlled mixed-evidence scenario that tests the model’s ability to selectively use relevant information while ignoring distractors. The dataset comprises 667 questions under this setting.
Answers are generated using LLaMA-3 8B Instruct with greedy decoding, and the resulting responses are evaluated by an external LLM judge, Gemini-2.5-Flash-Lite.

\label{sec:mix}
\paragraph{\textbf{Overall Pattern: Question-Guided Reasoning vs. External Over-Reliance}}

Across layers, a clear global contrast emerges:
\textbf{Correct predictions emphasize question understanding, whereas incorrect predictions over-rely on externally aligned answer content.}

As shown in Figure~\ref{fig: qq0}, correct samples consistently allocate more routing mass to \textbf{$Q \rightarrow Q$}. This indicates that the model places greater emphasis on understanding the question, first building a stable internal representation of the reasoning objective and then continuing to use it as a constraint while forming the answer.

Incorrect samples, in contrast, show relatively weaker question consolidation and stronger routing from \textbf{$Q \rightarrow \mathrm{Ans\_EXT}$}. This suggests a shortcut strategy: instead of constructing the answer through question-guided reasoning, the model leans heavily on answer fragments that are directly supported by retrieved content. As a result, the model tends to use context that is superficially aligned with the question, yielding locally plausible reasoning steps, while remaining  misaligned with the deeper semantic constraints required to correctly solve the problem.

Thus, the key difference is not simply how much external information is used, but \emph{whether answer formation is anchored in a well-formed question representation}.

\paragraph{\textbf{Layer-wise Distribution of Attribution Weight}}
While the previous section examined the overall routing distribution, we now provide a more detailed layer-wise analysis to better understand how routing patterns evolve across the network, as shown in Figure~\ref{fig: qq}  and Figure~\ref{fig: qin}.

\paragraph{Low Layers (0-7): Establishing the Question Anchor}

The divergence begins in the lowest layers. Correct predictions devote substantially more routing to $Q \rightarrow Q$, indicating stronger internal consolidation of the question before heavy involvement of answer-related representations. This early investment builds a stable semantic anchor that guides downstream reasoning.

Wrong predictions, however, show weaker $Q \rightarrow Q$ routing and relatively stronger early routing from $Q$ toward $\mathrm{Ans\_EXT}$. The model starts linking the question directly to externally aligned answer content before fully stabilizing the question representation itself. As a result, early processing is driven more by surface alignment with retrieved information than by a structured internal reasoning objective.

This early imbalance sets the stage for later errors: when question understanding is shallow, answer formation becomes vulnerable to external bias.





\paragraph{Higher Layers (8–31): Answer-Focused Refinement}

In the higher layers, routing patterns become broadly answer-centric for \emph{both} correct and wrong predictions. The dominant activity shifts toward refining internal answer representations, consistent with a late stage where the model mainly stabilizes the emerging answer and expresses it in fluent natural language. In this regime, the model primarily elaborates and stabilizes the evolving answer state. Correct predictions, however, still retain slightly stronger $Q \rightarrow Q$ routing than wrong ones, indicating a small but persistent influence of the question even at late stages.

\paragraph{\textbf{Summary: A Depth-Wise Shift from Question Guidance to External Drift}}

Correct and incorrect reasoning exhibit markedly different depth-wise trajectories. Correct predictions begin with strong comprehension of the question representation, followed by answer formation that remains consistently constrained by it—a pattern we term \textbf{question-constrained evidence grounding (QCEG)}. 
In contrast, incorrect predictions show weak early question grounding, transition prematurely toward externally aligned answer content, and ultimately refine answers under the dominance of externally driven. We refer to this failure mode as \textbf{surface-aligned evidence grounding (SAEG)}.

Thus, mixed-context errors arise from a progressive routing shift: computation moves away from question-guided reasoning and toward externally driven answer construction. Once this shift occurs in early layers, later processing may tends to amplify the misalignment.

\section{Graph-Structural Detection of Unfaithful Predictions}
\label{sec:method}

Our structural analysis reveals that correct and unfaithful predictions differ in the internal organization of their reasoning circuits. Correct answers tend to be supported by deeper, more integrated, and more coherent attribution graphs, whereas unfaithful answers arise from fragmented and weakly coordinated structures.

We leverage this insight by framing faithfulness detection as a \emph{graph classification} problem. If structural organization systematically differs between correct and unfaithful reasoning, then a model should be able to predict answer faithfulness directly from the structure of its attribution graph.

Concretely, given an attribution graph $G=(V,E)$ constructed from a model prediction, our goal is to estimate $p(y=1 \mid G)$, the probability that the prediction is faithful ($y=1$) rather than unfaithful ($y=0$), using only internal structural information. To achieve this, we employ a graph neural architecture that captures both local evidence propagation patterns and global circuit organization.

\subsection{Graph Features}






Each attribution graph $G$ contains nodes $v \in V$ representing question tokens, retrieved context tokens, intermediate activations, and generated tokens, and directed edges $(u,v) \in E$ representing causal attribution links. Each edge carries a scalar weight $w_{uv}$ indicating attribution strength.

\paragraph{Node features.}
Each node $v$ is associated with a feature vector $\mathbf{x}_v \in \mathbb{R}^{d_x}$, which concatenates a one-hot encoding of node type (e.g., question, context, answer, intermediate) with normalized structural signals such as in-degree, out-degree, total degree, and PageRank score. These features describe both the functional role and the local structural importance of each node.

\paragraph{Edge features.}
Each directed edge $(u,v)$ is assigned a one-dimensional feature $\mathbf{e}_{uv} = \tanh(w_{uv})$, a bounded transformation of the attribution weight. This scalar encodes how strongly information flows from $u$ to $v$ within the reasoning circuit.

\paragraph{Graph-level topology signatures.}
In addition to node- and edge-level information, we compute a vector of global structural statistics $\mathbf{g}(G) \in \mathbb{R}^{d_g}$, including measures such as longest-path depth, average degree, triad ratios, graph density, and maximum PageRank, as analyzed in Section~\ref{sec:graph-signatures} . These metrics summarize the overall organizational patterns of the circuit.

\subsection{Graph Transformer Encoder} To capture both local evidence propagation and long-range structural interactions, we use a  graph transformer encoder that alternates between message passing and attention. Let $\mathbf{h}_v^{(0)} = \mathrm{MLP}_{\text{in}}(\mathbf{x}_v)$ be the initial node embedding. Each layer $\ell = 1, \dots, L$ then performs two stages. \paragraph{Local structural propagation.} We first update node states using a message passing operator that aggregates information from immediate neighbors: \[ \tilde{\mathbf{h}}_v^{(\ell)} = \mathbf{h}_v^{(\ell-1)} + \mathrm{MPNN}\!\left(v,\{\mathbf{h}_u^{(\ell-1)}, \mathbf{e}_{uv}\}_{u \in \mathcal{N}(v)}\right), \] where $\mathcal{N}(v)$ denotes the neighbors of $v$. This step models how evidence locally accumulates along attribution edges. \paragraph{Global attention interaction.} We then apply a graph attention mechanism to allow non-local structural interactions: \[ \mathbf{h}_v^{(\ell)} = \tilde{\mathbf{h}}_v^{(\ell)} + \sum_{u \in V} \alpha_{vu}^{(\ell)} \mathbf{W}^{(\ell)} \tilde{\mathbf{h}}_u^{(\ell)}, \] where attention weights $\alpha_{vu}^{(\ell)}$ are computed from node features and edge features. This stage enables the model to capture long-range coordination across different parts of the reasoning circuit. After $L$ layers, each node has a final representation $\mathbf{h}_v^{(L)}$ encoding both local and global structural context.

\subsection{Graph-Level Readout}
\label{sec:readout}

We convert node embeddings into a graph representation using simple multi-statistic pooling:
\[
\mathbf{h}_{\text{pool}}(G)=
\big[\,\mathrm{mean}_{v\in V}\,\mathbf{h}_v^{(L)} \;\|\;
      \mathrm{sum}_{v\in V}\,\mathbf{h}_v^{(L)} \;\|\;
      \mathrm{max}_{v\in V}\,\mathbf{h}_v^{(L)}\,\big].
\]
Here $\|\,$ denotes concatenation. Mean pooling captures the overall structural tendency, sum pooling captures the total evidence mass, and max pooling captures the strongest pathway signal.

We additionally embed a small set of global topology statistics $\mathbf{g}(G)\in\mathbb{R}^{d_g}$:
\[
\mathbf{h}_g(G)=\phi_g(\mathbf{g}(G)),
\]
where $\phi_g$ is a small MLP. The final graph representation is simply
\[
\mathbf{z}(G)=\big[\mathbf{h}_{\text{pool}}(G)\;\|\;\mathbf{h}_g(G)\big].
\]
A final classifier produces $p(y=1\mid G)$ from $\mathbf{z}(G)$.

\subsection{Unfaithful Prediction}

A final classifier produces the probability that the prediction is correct:
\[
p(y=1 \mid G) = \mathrm{softmax}\big(\mathrm{MLP}_{\text{out}}(\mathbf{z}(G))\big)_1.
\]
We predict an output as unfaithful when $p(y=1 \mid G) < 0.5$.

This detector checks whether the model’s internal reasoning forms a coherent structure or a fragmented one.
Local message passing tracks how evidence flows along attribution paths, global attention connects distant but related parts of the circuit, and graph-level statistics summarize the overall depth and organization.
Together, these signals reveal reasoning failures that are not visible from the answer text or retrieved documents alone.

\subsection{Results}

We now evaluate whether internal reasoning structure can be used to reliably predict when a RAG model’s answer is correct. 

\subsubsection{Experimental Setup.}
We evaluate detection on attribution graphs derived from HotpotQA, 2WikiMultihopQA, and MuSiQue. For each dataset, we follow the fixed graph-based split defined during graph construction: up to 500 wrong and 500 correct examples are collected in deterministic filename order, 250 per class are sampled for training/validation, and the remaining 500 examples form a balanced test set. All methods are evaluated on the same fixed indices to ensure direct comparability.

As a non-structural baseline, we use a logit-based self-judging signal computed from the model’s own output distribution, without access to gold answers. After generating an answer, the same model is prompted to judge whether its prediction is correct given the question, retrieved context, and its own reasoning trace, and is restricted to a binary response (``Yes'' or ``No''). We compute $\log p(\text{Yes})$ and $\log p(\text{No})$ for the two continuations and predict correctness when $\log p(\text{Yes}) > \log p(\text{No})$. 

\subsubsection{Graph Detector Training.}
The graph detector uses a graph transformer encoder with $L=2$ layer and hidden size 128, trained with AdamW (learning rate $10^{-4}$, batch size 32) and dropout 0.1. 
For each dataset, we construct a balanced set of 1{,}000 graphs (500 incorrect and 500 correct) and use a fixed split with 250 graphs per class for training/validation and 250 per class for testing.

\subsubsection{Detection Performance.}
\begin{figure}     \centering     
\setlength{\abovecaptionskip}{0.cm}     \setlength{\belowcaptionskip}{0.cm}     \includegraphics[scale=0.27]{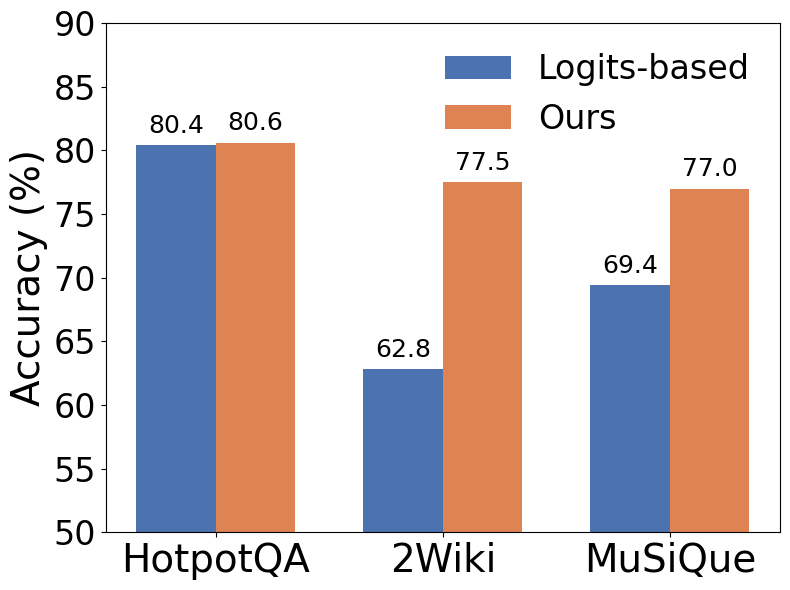}  
\captionsetup{skip=10pt}
\vskip -1em
\caption{Performance comparison across QA benchmarks.
}   \label{fig: three} 
\vskip -2em
\end{figure}

Figure~\ref{fig: three} shows that the graph-structural detector consistently outperforms the logit-based self-judging baseline on all three QA benchmarks. Averaged across datasets, our method improves accuracy by 11.53\%.

These results demonstrate that modeling internal reasoning structure provides a substantially more reliable signal of answer correctness than relying on the model’s own output confidence. While logit-based self-evaluation reflects surface-level uncertainty, it cannot determine whether evidence has been integrated through a coherent multi-step reasoning circuit. In contrast, the graph-structural detector directly measures the organization of evidence flow, enabling more accurate detection of unfaithful predictions under challenging retrieval conditions.


\section{Attention Intervention for RAG Improvement}
\label{sec:intervention}

Section~\ref{sec:mix} shows that incorrect predictions under mixed retrieval are not random errors but arise from a routing pattern: the model under-invests in early question consolidation, over-commits to surface-matched context, and later drifts into self-reinforcing decoding that is weakly constrained by the question. This section turns that diagnosis into an actionable control mechanism. Instead of changing parameters or re-retrieving, we intervene directly on the attention computation during decoding to encourage the routing behavior characteristic of correct predictions.

\subsection{Intervention as Layer-Wise Routing Control}

The analysis above reveals that RAG errors arise from a systematic shift in how information is routed across layers. 
Correct predictions maintain strong question grounding throughout the computation, whereas wrong predictions exhibit two coupled failures: (i) insufficient early consolidation of the question representation, and (ii) progressively increasing reliance on externally information.

We therefore design an intervention that directly reshapes routing preferences inside the attention mechanism. The goal is not to retrain the model, but to gently bias information flow toward the routing regime associated with correct reasoning.
We partition token positions into three semantically meaningful regions:

\begin{itemize}
    \item $Q$: question tokens,
    \item $Ex$: external retrieved context tokens,
    \item $In$: internally generated tokens.
\end{itemize}







\subsubsection{Control 1: Strengthening Early Question Understanding}

Section~\ref{sec:mix} shows that wrong predictions underutilize $Q \rightarrow Q$ routing in early layers. As a result, the question representation is not sufficiently understood before interacting with retrieved context.

To counteract this, we amplify attention among question tokens in lower layers:
\[
\alpha^{(\ell)}_{Q \rightarrow Q} = \alpha_{\text{QQ}} \quad \text{for } \ell \in \mathcal{L}_{\text{low}}, \quad \alpha_{\text{QQ}} > 1.
\]

This encourages deeper internal integration of question semantics before heavy evidence mixing occurs.

\subsubsection{Control 2: Suppressing Premature Context Reliance}

We further observe that incorrect predictions tend to route attention toward external context too early, leading to brittle surface alignment rather than deep reasoning.

We therefore down-weight attention whose *target* lies in the external region during the same early stage:
\[
\alpha^{(\ell)}_{\ast \rightarrow Ex} = \alpha_{\text{ctx}} 
\quad \text{for } \ell \in \mathcal{L}_{\text{low}}, \quad \alpha_{\text{ctx}} < 1.
\]

Here $\ast$ denotes any source region.
This control reduces the influence of retrieved tokens before the question representation has stabilized.

\subsubsection{Control 3: Maintaining Question-Guided Decoding}

In later layers, answer tokens are iteratively refined. For correct predictions, routing from the question to internal answer states remains active, ensuring that decoding stays constrained by the original task. Incorrect predictions, in contrast, show weakening $Q \rightarrow In$ routing and increasing self-reinforcement among answer tokens.

To maintain question guidance during decoding, we strengthen attention from question tokens to internally generated tokens in higher layers:
\[
\alpha^{(\ell)}_{Q \rightarrow In} = \alpha_{\text{QIn}} 
\quad \text{for } \ell \in \mathcal{L}_{\text{high}}, \quad \alpha_{\text{QIn}} > 1.
\]


\subsubsection{How is the Control Applied in the Model?}

The proposed intervention does not alter any model parameters and requires no retraining. Instead, it operates directly on the model’s internal attention computation during inference. We implement this control using forward hooks inserted into selected transformer layers.

A hook is a lightweight function inserted into the model’s forward pass that intercepts and modifies intermediate activations without changing model parameters. In our case, the hook intercepts the attention pattern immediately before it is used to aggregate value vectors. At chosen layers, we rescale specific groups of attention weights based on the semantic regions of the source and target tokens (question, retrieved context, or generated answer tokens) as well as the layer index.

Because the modification occurs at the level of attention weights, it changes the relative influence that different token groups exert on one another, thereby steering information routing inside the network. Importantly, all original model weights remain frozen: the intervention introduces only a small amount of element-wise scaling applied on-the-fly during the forward pass. As a result, the computational overhead is negligible, and the model’s base behavior can be fully restored by simply removing the hooks.






\subsection{Results}
\begin{figure}     \centering     
\setlength{\abovecaptionskip}{0.cm}     \setlength{\belowcaptionskip}{0.cm}     \includegraphics[scale=0.31]{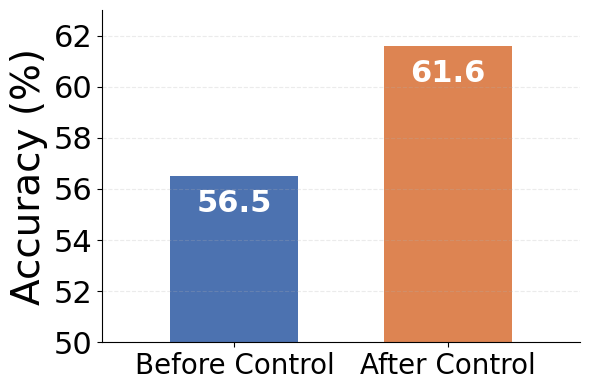}  
\captionsetup{skip=10pt}
\vskip -1em
\caption{Performance comparison on Mix-MusiQue.
}   \label{fig:mix} 
\vskip -1em
\end{figure}

We now evaluate whether the proposed layer-wise routing control improves answer faithfulness in the mixed-context setting. We focus on overall answer accuracy, which reflects whether the model ultimately produces the correct answer under the presence of both supporting and distracting evidence.

\subsubsection{Setup}

Our experiments focuses on a mixed-context setting that we construct based on MuSiQue, which we refer to as \textbf{Mix-MuSiQue}. In this dataset, each question is paired with a retrieved context that intentionally contains both supporting and non-supporting passages. The final evaluation set consists of 667 questions under this mixed-evidence condition.
Answers are generated using LLaMA-3 8B Instruct. We then evaluate the responses using an external LLM judge, Gemini-2.5-Flash-Lite.

We compare two conditions: Before Control, the standard model without intervention, and After Control, where region-aware attention reweighting is enabled in lower and higher layers. All other decoding settings are kept identical to ensure a fair comparison. For our method, we set $\alpha_{\text{QQ}} = 1.5$, which promotes stronger understanding of question semantics before substantial interaction with retrieved evidence. We set $\alpha_{\text{ctx}} = 0.5$, mitigating premature over-reliance on external information in early layers. Finally, we set $\alpha_{\text{QIn}} = 1.5$, which biases later computation toward refining answer representations under continued guidance from the question.

\subsubsection{Intervention Results in the Mixed-Context Setting}
Figure~\ref{fig:mix} reports performance on the Mix-Musique setting, where the retrieved context contains both supporting and non-supporting information. This mixed-evidence scenario is particularly challenging because the model must distinguish useful evidence from distractors while maintaining alignment with the question.

Without intervention, the baseline model attains 56.5\% accuracy. Applying our attention control raises performance to 61.6\%, representing a 9\% improvement. This consistent gain demonstrates that the proposed intervention effectively reshapes the model’s internal information routing under mixed-context conditions.

The improvement suggests that errors in this setting  are closely tied to how the model allocates attention across question and context tokens. By strengthening question-grounded routing and reducing early over-reliance on retrieved context, the intervention helps the model better integrate relevant evidence while suppressing distractors, leading to more faithful answer generation. 
In addition, we present a case study in Appendix~\ref{sec:case} due to space limitations.

\section{Conclusion}

We present a graph-based perspective on why retrieval-augmented generation can fail even when relevant evidence is available. Attribution graphs reveal clear structural differences between success and failure: correct predictions show deeper, more distributed, and question-constrained evidence integration, while errors exhibit shallow, context-dominated routing. These insights enable graph-based error detection and targeted inference-time interventions that reshape internal information flow. 





\newpage
\bibliographystyle{ACM-Reference-Format}
\bibliography{sample-base}

\newpage
\appendix

\section{Appendix}

\subsection{Related Work}
\label{sec:related}
\textbf{Retrieval-Augmented Generation.}
Retrieval-Augmented Generation (RAG) has become a widely adopted paradigm for improving the factuality and reasoning capabilities of large language models by grounding generation in external knowledge sources~\cite{zhao2026retrieval,fan2024survey}. Prior work has explored a broad range of retrieval strategies, including dense and hybrid retrievers~\cite{nian2025w}, multi-hop retrieval~\cite{tang2024multihop}, iterative retrieval–generation loops~\cite{trivedi2023interleaving}, and query reformulation~\cite{chan2024rq}. These efforts have demonstrated that retrieval quality and evidence coverage play a critical role in downstream performance, and have led to significant gains on question answering and knowledge-intensive benchmarks.

More recent studies have focused on improving RAG robustness through better context selection, reranking, compression, or prompt engineering~\cite{dong2024don,ampazis2024improving}. While these methods reduce failures at the system level, they primarily treat the language model as a black box and do not address how retrieved evidence is internally processed once injected into the context. As a result, they offer limited insight into why errors persist even when relevant evidence is successfully retrieved. 

In the context of RAG, several works have proposed methods to assess faithfulness, evidence usage, and citation correctness~\cite{zeng2025towards,liu2025selfelicit}. However, these approaches typically rely on representations from a single layer and provide only a largely static view of the model’s internal state, lacking a comprehensive and dynamic perspective for analyzing how evidence is integrated during computation.

\textbf{Interpretability and Circuit Analysis of LLMs} 
Parallel to advances in RAG, a growing body of work has investigated the internal mechanisms of transformer models using interpretability techniques such as attention analysis~\cite{clark2019does,vig2019analyzing}, Sparse Autoencoders (SAEs)~\cite{cunningham2023sparse}, transcoders~\cite{dunefsky2024transcoders}, and circuit tracing~\cite{elhage2021mathematical}. These studies suggest that specific reasoning behaviors can often be traced to localized subnetworks or circuits spanning multiple layers and attention heads.
A common theme across these methods is the decomposition of dense neural representations into more interpretable feature spaces~\cite{paulo2024automatically,ferrando2024know}, which in turn enables the construction of attribution graphs that model the causal flow of information within the network~\cite{marks2024sparse}.

While circuit-level analyses have provided valuable insights into question answering~\cite{dai2025graphghost,zhao2025verifying} by verifying chain-of-thought reasoning through computational graphs, most prior work has been limited to standalone language models without retrieval. As a result, the interaction between external evidence and internal circuits in RAG settings remains insufficiently understood.

\textbf{Positioning of Our Work.}
Our work bridges these lines of research by providing a mechanistic, circuit-level analysis of RAG systems. Unlike prior RAG studies that emphasize retrieval quality or prompt design, we focus on how retrieved evidence propagates through internal model circuits. By introducing attribution graphs, we move beyond scalar attribution scores and enable structural analysis of information flow between faithful and hallucinated generations. This perspective allows us to identify integration failures inside the model as a key source of RAG failures, complementing and extending existing system-level analyses.
\subsection{Case Study}
\label{sec:case}
Figure~\ref{fig: case} provides representative examples illustrating how the intervention improves reasoning under mixed-context conditions. In both examples, the baseline model fails to complete the full reasoning chain: in one case it stops at a salient intermediate entity without performing the final role mapping, and in the other it fails to bridge a geographic clue to the relevant historical event. After attention control is applied, the model follows a more complete reasoning path, successfully connects intermediate evidence, and produces the correct final answer.

These qualitative examples show that the intervention does not merely alter surface-level outputs. Instead, it changes how the model integrates evidence across reasoning steps, improving its ability to link intermediate clues to the final answer.
\begin{figure}[h]     \centering     
\setlength{\abovecaptionskip}{0.cm}     \setlength{\belowcaptionskip}{0.cm}     \includegraphics[scale=0.19]{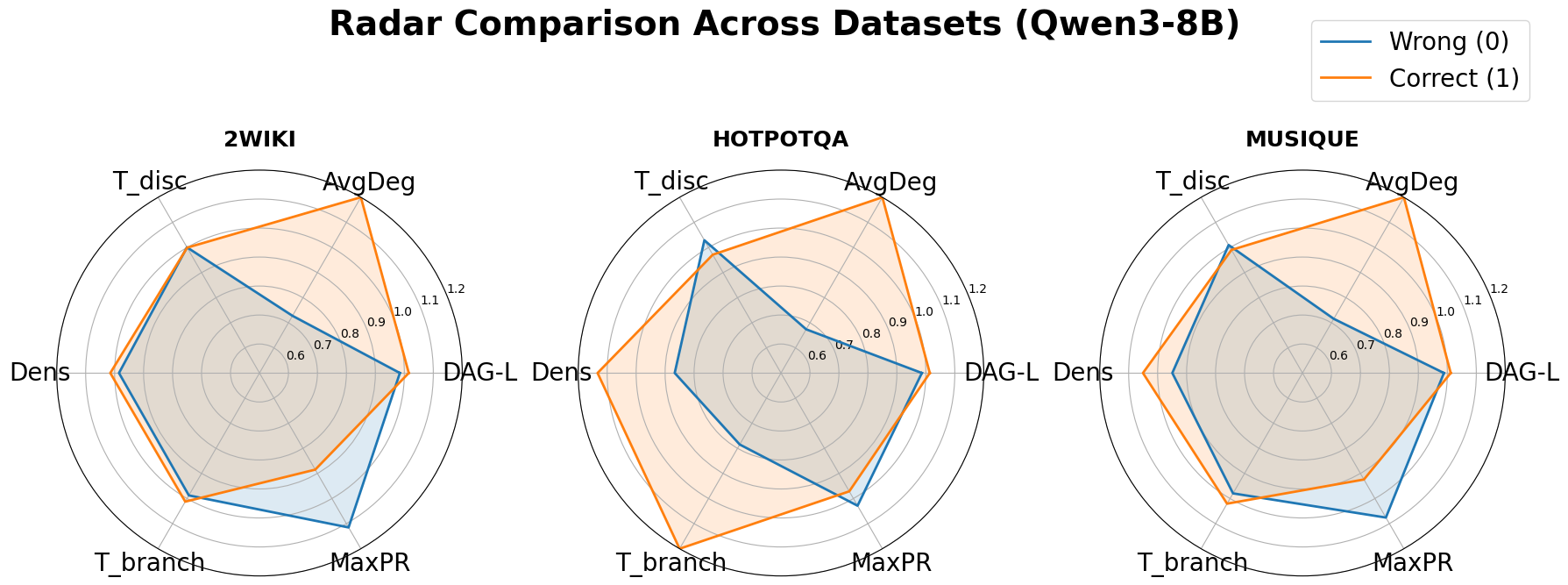}  
\captionsetup{skip=10pt}
\vskip -1em
\caption{Radar comparison of  attribution-graph structural metrics between correct and incorrect predictions across three QA datasets (2Wiki, HotpotQA, and MuSiQue).}   \label{fig:qwen} 
\vskip -0.4em
\end{figure}

\begin{figure}     \centering     
\setlength{\abovecaptionskip}{0.cm}     \setlength{\belowcaptionskip}{0.cm}     \includegraphics[scale=0.17]{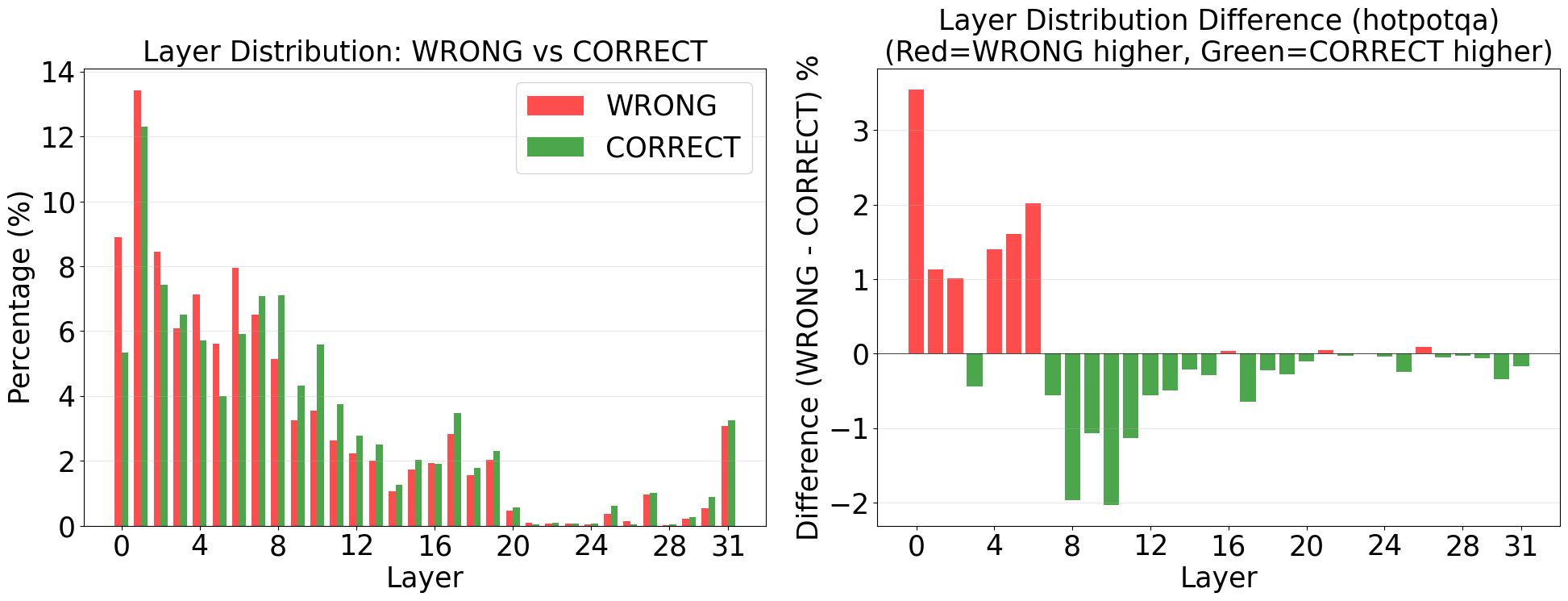}  
\captionsetup{skip=10pt}
\vskip -1em
\caption{Layer-wise attribution mass for correct and wrong predictions (left) and their difference on HotpotQA (right).
}   \label{fig: shallow_hot} 
\vskip -0.4em
\end{figure}

\begin{figure}     \centering     
\setlength{\abovecaptionskip}{0.cm}     \setlength{\belowcaptionskip}{0.cm}     \includegraphics[scale=0.17]{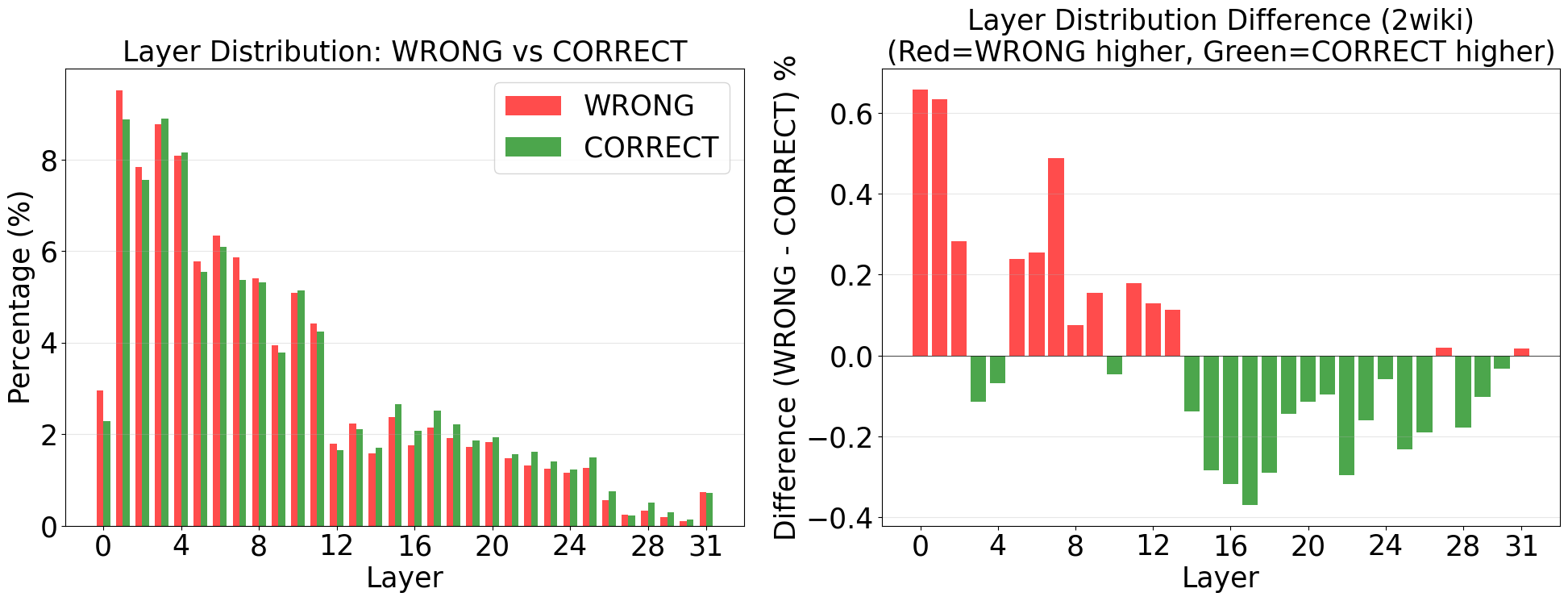}  
\captionsetup{skip=10pt}
\vskip -1em
\caption{Layer-wise attribution mass for correct and wrong predictions (left) and their difference on 2Wiki (right).
}   \label{fig: shallow_2wiki} 
\vskip -0.4em
\end{figure}
\begin{figure*}[t]
\centering

\begin{tabular}{|p{3cm}|p{12cm}|}
\hline
\multicolumn{2}{|l|}{\textbf{Successful Intervention Case 1}} \\ \hline

\textbf{Question} &
Who does the singer of May We All with Florida Georgia Line play in Country Strong? \\ \hline

\textbf{Gold Answer} & James Canter \\ \hline

\textbf{Baseline Model Output (Before Control)} &
The singer of ``May We All'' with Florida Georgia Line is Tim McGraw.  
He appears in \textit{Country Strong}.  
\textbf{Final answer: Tim McGraw.} \\ \hline

\textbf{Controlled Model Output (After Control)} &
The singer of ``May We All'' with Florida Georgia Line is Tim McGraw.  
In \textit{Country Strong}, Tim McGraw plays the role of James Canter.  
\textbf{Final answer: James Canter.} \\ \hline

\textbf{What Changed} &
The baseline system stopped at the intermediate entity (the actor) and failed to perform the final role-mapping step.  
After intervention, the model completed the reasoning chain and produced the correct character name instead of the actor’s name. \\ \hline

\end{tabular}

\vspace{0.8em}

\begin{tabular}{|p{3cm}|p{12cm}|}
\hline
\multicolumn{2}{|l|}{\textbf{Successful Intervention Case 2}} \\ \hline

\textbf{Question} &
When did allied troops land in the region sharing a border with Sahel? \\ \hline

\textbf{Gold Answer} & 8 November 1942 \\ \hline

\textbf{Baseline Model Output (Before Control)} &
The context mentions North Africa and Operation Torch,  
but it does not explicitly state the landing date for allied troops in the region bordering Sahel.  
\textbf{Final answer: Not mentioned.} \\ \hline

\textbf{Controlled Model Output (After Control)} &
The region sharing a border with Sahel refers to North Africa.  
Operation Torch was the Allied invasion of French North Africa,  
which began on 8 November 1942.  
\textbf{Final answer: 8 November 1942.} \\ \hline

\textbf{What Changed} &
The baseline system failed to link the geographic clue (Sahel) to North Africa and therefore missed the historical event.  
After intervention, the model successfully bridged the geographic reference to Operation Torch and retrieved the correct date. \\ \hline

\end{tabular}

\caption{
Examples where routing control improves multi-hop reasoning.  
In both cases, the baseline model either stops at an intermediate entity or fails to bridge a geographic clue to a historical event.  
After control, the model follows a more complete reasoning path and produces the correct answer.
}
\label{fig: case}

\end{figure*}

\end{document}